# Causal Machine Learning in IoT-based Engineering Problems: A Tool Comparison in the Case of Household Energy Consumption

N.-L. Kosioris[1,2]    S. Nikoletseas[1]    G. Filios[1]    S. Panagiotou[1]

[1] Department of Computer Engineering and Informatics, University of Patras

[2] Connect Line S.A., Athens, Greece

**Abstract:**   The rapid increase in computing power and the ability to store Big Data in the infrastructure has enabled predictions in a large variety of domains by Machine Learning. However, in many cases, existing Machine Learning tools are considered insufficient or incorrect since they exploit only probabilistic dependencies rather than inference logic. Causal Machine Learning methods seem to close this gap. In this paper, two prevalent tools based on Causal Machine Learning methods are compared, as well as their mathematical underpinning background. The operation of the tools is demonstrated by examining their response to 18 queries, based on the IDEAL Household Energy Dataset, published by the University of Edinburgh. First, it was important to evaluate the causal relations assumption that allowed the use of this approach; this was based on the preexisting scientific knowledge of the domain and was implemented by use of the in-built validation tools. Results were encouraging and may easily be extended to other domains.

**Abbreviations**

| | |
|---|---|
| ML | Machine Learning |
| CML | Causal Machine Learning |
| IML | Interpretable Machine Learning |
| EML | Explainable Machine Learning |
| CSM | Structural Causal Model |
| ATE | Average Treatment Effect |
| ITE | Individual Treatment Effect |
| MSE | Mean Squared Error |
| MLP | Multi-Layer Perceptron |
| KLD | Kullback-Leibler Divergence |
| AUUC | Area Under Uplift Curve |

## 1   Introduction

The majority of ML algorithms used today have the ability to detect patterns of any scale in a wide range of datasets from small to huge. However, some of the resulting data-driven models suffer from biases, difficulty in generalizing for future uses and overfitting, as well as a lack of distinction between probabilistic association and true causation (Mehrabi et al., 2021). Human control in this context is often impossible to achieve due to complexity, resulting in erroneous conclusions. From the users' point of view such models are no longer sufficient, as results need to be explained in a cause-and-effect format, which is closer to the human way of thinking (Schölkopf, 2022). In this way, the effect of changes in the underlying assumptions of the model is easier to understand, saving time from re-creating a dataset. Incorporating causality logic into existing AI systems should handle the above problems making models more efficient and intelligent, reducing the associated energy- and time-consuming learning cycles for processes that are trivial for humans. Thus, CML tools are expected to play a leading role in the coming years.

CML is a major milestone in ML that enables AI models to make accurate predictions based on causes rather than just probabilistic correlations, thereby allowing the understanding of their effect. Note that causal hypotheses are not already included in the statistical description of the system. Some common pitfalls that Machine Learning falls into and that CML focuses on solving are listed below:

- *Correlation Equation with Causation.* Predictive algorithms can fall into the trap of equating correlation with causation, i.e. believing that because event X precedes event Y, it must be the cause of it.
- *Inability to use non-independent and identically distributed random variables.* ML mostly implement pattern recognition in large data sets for which random observations are independent of each other and occurrence probability is continuous. However, in real-world scenarios, data distribution can change, e.g. in object recognition from images under unusual/novel lighting, angles or backgrounds (Cheng et al., 2022).
- *Absence of Immutability.* Current ML algorithms are limited in their ability to generalize the patterns they find in a training dataset for multiple applications (G. Xu et al., 2020). Ideally, the ML model should exhibit immutability, i.e. generalize beyond the limited set of environments that can be accessed for training, and even extend to new and perhaps unpredictable environments.
- *Lack of Explainability.* ML models remain mostly "black boxes" that cannot explain the reasons behind their predictions or recommendations (Kaddour et al., 2022). Typically, Deep Learning systems can be trained to recognize manufacturing process defect images with high accuracy, but - unlike a real expert - it cannot explain why or how a specific image indicates a defect.

Interestingly, CML, Interpretable Machine Learning (IML) and Explainable Machine Learning (EML) are in a sense related, even complementary. CML is aiming to understand what will happen as a result of an action, i.e. it focuses on cause-and-effect, whereas IML aims to understand how the model functions, focusing on transparency, explanation, and trust. The relevant tools will, as a result, also differ. Furthermore, EML differs from IML in that it provides methods that help explain complex, black-box models, which are not inherently interpretable, so post-hoc tools are needed to explain their behaviour. By contrast, IML refers to models that are inherently understandable by humans without extra tools (Linardatos et al., 2021), (G. Xu et al., 2020)

The aim of this work is to highlight and compare state-of-the-art tools in the field of CML by a sample application in an IoT-data environment, highlighting ways to exploit the resulting model. Furthermore, the extended aim of the work is to guide future adaptation of the tools to other domains and use cases.

The paper is structured in the following way: In Section 2, published work pertaining to this paper is briefly reviewed. In Section 3, the notional foundation of the architectures that have prevailed so far in CML are presented. Then, in Section 4, the two most popular and powerful tools in the field are presented, namely Microsoft PyWhy$^{TM}$ and UberML$^{TM}$, as well as the rest of the tools needed in implementing CML models. Further, in Section 5, the selected data set is presented and its required pre-processing is clarified. In Section 6, the models that resulted from the two tools are represented graphically, followed by an explanation of the assumptions regarding the data set and the methods of testing and evaluating the results. Section 7 summarizes conclusions and proposes future extension directions of this work.

## 2 Related Work

The adoption of CML for engineering problems aims to provide a framework moving beyond correlation-based insights toward cause-effect modelling. In particular, in cases related to data stemming from IoT, CML methods provide a foundation for interpreting sensor data, designing interventions, and optimizing the function of the pertinent engineering system. (Naser, 2021).

The broader application topic of this paper concerning energy management is probably the one attracting most registered causal AI approaches. For instance, (Massidda & Marrocu, 2023) propose a novel method for probabilistic forecasting of the total load of a residential community and its base and thermal components, combining conformalized quantile regression and causal machine learning, using only aggregate consumption and environmental conditions in a dataset of a residential community in Germany. The T-learner method was the most effective among the causal methods for load disaggregation in terms of accuracy, simplicity, and potential for extension.

(C. R. Kinttel & Stolper, 2019) use causal forests to evaluate the heterogeneous treatment effects (TEs) of repeated behavioral nudges towards household energy conservation. Pre-treatment consumption and home value are the strongest predictors of treatment effect. A "boomerang effect" is noticed: households with lower consumption than similar neighbours are the ones with positive TE estimates.

(Miraki et al., 2024) propose eXplainable Causal Graph Neural Network (X-CGNN) for multivariate electricity demand forecasting yielding intrinsic and global explanations based on causal inferences as well as local explanations based on post-hoc analyses. State-of-the-art performance was attained with two real-world electricity demand datasets from both the household and distribution levels.

In (Duhirwe et al., 2024) a causal inferential approach is employed combining double machine learning and domain knowledge using directed acyclic graphs. The 2015 US Residential energy consumption survey is used. The results show that energy audits, proper insulation, access to interval data etc. significantly reduce energy use intensity (EUI). For instance, proper insulation reduces EUI by 5.6 MWh/m2 while changing static thermostat settings to automatic adjustments at certain times reduces EUI by 3.5 MWh/m2.

(Kim & Park, 2025) implemented a causal modeling approach using Bayesian Networks (BNs), focusing on inter-causalities among various building occupant behaviors. Data were collected from a yearlong study involving six households in Seoul, South Korea. In the customized BN model, the intercausal relationships for each household were quantitatively measured using machine learning, revealing distinct personal preferences and interactions with the environment and high predictive accuracy.

Data-driven building thermal dynamics models circumvent the need for domain experts and determination of physical properties of the buildings, but often fail to generalize to truly unseen conditions since they are not causal. (Jiang & Kazmi, 2025) demonstrate that causal machine learning (CML) algorithms trained on debiased data from nine real-world Dutch buildings can produce accurate models necessary for control-oriented applications which outperform baseline models by over 40%, besides learning the correct causal associations which were verified using a custom testing environment and SHAP feature analysis.

Taking an example from energy-efficient building design (Chen et al., 2022) introduce causal inference proposing a four-step process to discover and analyze parametric dependencies in a mathematically rigorous and computationally efficient manner by identifying the causal diagram, which provides a nexus for integrating domain knowledge with data-driven methods, towards interpretability and testability against the domain experience within the design space.

(Maisonnave et al., 2024) present an alternative approach to predictive modelling for future energy demands based on an ensemble causal model generated by four prominent causal detection methods using a dataset of the wholesale electricity company of Argentina (CAMMESA).

(Shams Amiri et al., 2023) demonstrate an Extreme Gradient Boosting (XGBoost) algorithm for forecasting the energy use of commercial and residential buildings in Philadelphia, USA. A corresponding SHAP (SHapley Additive exPlanations) analysis is implemented to pinpoint feature contributions to the model's energy estimates. By using the PopGen software, the model's energy estimates could be analyzed at the household level, the smallest possible scale. The results indicate that features related to lower building intensity (e.g., lower square footage, fewer floors per building) were associated with reduced energy use, whereas ''single-family attached'' zoning designation corresponds with higher energy estimates.

Energy retrofits of existing buildings provide an effective means to reduce building consumption and carbon footprints currently accounting for 40% of the energy consumption and 31% of the $CO_2$ emissions in US. (Y. Xu et al., 2021) uses data from a portfolio of 550 federal buildings exploiting the causal forest estimator to predict the retrofit effect as a function of building characteristics, climate, energy-use level, etc. A causal forest is an ensemble of many causal trees. A causal tree was fitted for each type of retrofit action and fuel type (electricity and gas) and was built with a random subsample of the whole data set. Within the findings, a dashboard tool and fault detection system, commissioning, and HVAC investments had the highest average savings among the six actions analyzed.

Demand response (DR) is pivotal in enhancing the operational efficiency and reliability of power systems. (He & Khorsand, 2024) integrate causal learning approaches for causal intervention and counterfactual analyses of prosumers' DR participation and consumption behavior. The proposed framework incorporates domain knowledge of the power system to enhance model accuracy and performance, demonstrated by simulation.

(Sun et al., 2024) reviews the application of causal relationships in power systems fault diagnosis, energy prediction, stability analysis, and the electricity market, looking forward to the development prospects of causal relationships in combination with power physics models and artificial intelligence, towards reliable and intelligent new power systems focusing on renewable sources.

Widening the engineering perspective, diverse problems have been tackled with causal AI mostly in the domains of robotics/automation and manufacturing.

In robotics, (Castri et al., 2023) propose an extension of a state-of-the-art causal discovery method embedding an additional feature-selection module based on transfer entropy. Starting from a prefixed set of variables, the new algorithm reconstructs the causal model of the observed system by considering only its main features and neglecting those deemed unnecessary for understanding the evolution of the system. The method is validated on a real-world robotics scenario using a large-scale time-series dataset of human trajectories. (Hellström, 2021) investigate the role of causal reasoning in robotics research. Inspired by a categorization of human causal cognition, a categorization of robot causal cognition is proposed mainly covering the sense–plan–act level of robotics, but also understandability, machine ethics, and robotics research methodology. (Nadim et al., 2023) propose an intelligent supervisory control approach which adopts deep reinforcement learning (DRL) to develop an efficient control policy through interaction with a process simulation. The DRL training history is then exploited using interpretable machine learning and process mining to build a discrete event system (DES) model, in the form of a state-event graph. The DES model identifies causal relationships between events and provides interpretability to the control policy developed by the DRL method.

In the manufacturing domain, (Vuković & Thalmann, 2022) provide a comprehensive overview about how causal discovery can be applied identifying four core areas, namely fault detection, analysis and management, root cause analysis, causality in a facilitator role, and domain and conceptual work. In (Ko et al., 2023), a novel physics-guided data-driven framework for Additive Manufacturing includes a Process-Structure-Property (PSP) causal relationships learning process with two sub-processes, namely a knowledge-graph-guided top-down approach to generate the requirements for predictive analytics and data acquisition and a data-driven bottom-up approach to construct and model new PSP knowledge. (Hua et al., 2022) proposed a zero-shot prediction method for cutting tool wear prediction based on causal inference. A deep convolutional neural network and a causal representation model are jointly optimized to extract invariant causal signal features, which can be generalized to non-stationary manufacturing environments without any new data. In (Wu & Wang, 2021) cause–effect relations intrinsic to an engineering design problem are employed to decompose a pertinent simulation-based causal ANN into sub-networks. Attractive subspaces are identified from the causal-ANN by leveraging its structure and the theory of Bayesian Networks. In (Zhou et al., 2024) a causal quality-related knowledge graph (CQKG) is constructed regarding quality defects in aerospace product manufacturing followed by a quality-related prompt dataset with multi-round conversations. Then, a novel P-tuning that adapts to utilize external CQKG instructions is designed to fine-tune an open-source ChatGLM base model. Based on this, a causal knowledge graph augmented LLM, named CausalKGPT, is developed assisting workers to analyze quality defects.

In summary, CML has demonstrated its value across multiple engineering domains, consistently outperforming traditional methods when interpretability, counterfactual reasoning, and intervention design are required. The reviewed studies employed a variety of computational tools. The same tools applied to IoT-based household energy systems enables fine-grained causal modelling of device interactions, occupancy behavior, and external influences. This sets the stage for developing intelligent, explainable systems capable of adaptive energy management, targeted fault detection, and personalized interventions.

# 3 CML Foundations

The mathematical foundations of CML draw from causal inference theory, which focuses on estimating the effects of interventions or changes in one variable on another. CML builds upon several foundational concepts, such as Directed Acyclic Graphs, counterfactuals, and do-calculus.

A primary mathematical tool in causal inference is "Directed Acyclic Graphs (DAGs)". In a DAG, nodes represent variables, and directed edges (arrows) represent causal relationships between those variables. A directed edge from variable $X$ to variable $Y$ indicates that $X$ has a causal effect on $Y$. The acyclic nature of the graph ensures that no variable can causally influence itself, either directly or indirectly, which reflects the directionality of cause-and-effect. Mathematically, a DAG can be used to represent the structure of a causal system, where each node in the graph is associated with a random variable, and edges indicate dependencies. This structure allows us to distinguish between direct and indirect causal effects, which is crucial for understanding how variables influence each other in a system.

A core concept in causal inference is the "Potential Outcomes Framework", also known as the "Rubin Causal Model" (RCM). This framework defines the causal effect of a treatment or intervention in terms of potential outcomes. Suppose we are interested in the effect of treatment $T$ on outcome $Y$. Each individual has two potential outcomes: $Y(1)$, the outcome if the individual receives the treatment, and $Y(0)$, the outcome if they do not. The causal effect for an individual is the difference between these two potential outcomes: Causal Effect $= Y(1) - Y(0)$. However, in practice, we can only observe one of these outcomes for any individual, depending on whether they receive the treatment. This is known as the "fundamental problem of causal inference".

To estimate causal effects, we rely on techniques like "propensity score matching", where we compare individuals with similar characteristics but different treatments, and "instrumental variables" (IV), which help address confounding factors that may bias the estimation of causal effects.

The framework of "do-calculus", introduced by Judea Pearl, is another powerful tool in causal inference. The do-operator, denoted as $do(X = x)$, represents an intervention that sets the variable $X$ to a specific value $x$, effectively "forcing" the system to adopt this value, independent of its natural distribution. The do-calculus provides a set of rules for manipulating causal graphs to compute the effects of interventions. For example, if we want to compute the effect of an intervention $do(X = x)$ on the outcome $Y$, we can use the do-calculus to break down the problem into simpler subproblems, manipulating the graphical model to identify the relevant causal paths and avoid confounding influences. In a simplified sense, this process helps us estimate the expected outcome of an intervention, accounting for the causal relationships present in the system.

In real-world scenarios, causal effects are often estimated from "observational data" rather than controlled experiments. One of the key challenges is dealing with confounding, where a third variable influences both the treatment and the outcome. Techniques like "propensity score matching" and "instrumental variables" are used to estimate causal effects in the presence of confounders. For instance, in propensity score matching, we estimate the likelihood of receiving a treatment based on observed covariates, and then match treated and untreated individuals with similar propensity scores. This helps control for confounding and provides an unbiased estimate of the treatment effect.

In modern machine learning, causal reasoning is integrated into algorithms to learn not just correlations but also causal relationships. "Causal Bayesian Networks" is an example of how machine learning models can be extended to learn causal structures from data. These models rely on the principles of DAGs and the potential outcomes framework to incorporate causal relationships into predictive models.

# 4 CML Implementation Tools

## 4.1 Causal Inference Tools

In recent years, the integration of causal inference with machine learning has gained significant traction, driving the development of tools such as PyWhy[TM] and CausalML[TM]. Both are Python libraries designed to facilitate causal analysis, but they differ in scope, methodology, and application domains.

### 4.1.1 Capabiltites

PyWhy[TM], developed by Microsoft, emphasizes a principled framework for causal inference. It is grounded in Judea Pearl's do-calculus and supports the entire causal inference pipeline: defining a causal graph, identifying causal effects, estimating them, and refuting the estimates. PyWhy[TM] primarily supports individual treatment effect (ITE) estimation using methods like propensity score matching and stratification. It allows users to incorporate domain knowledge explicitly via causal graphs and focuses on rigorous assumptions checking and refutation testing, making it suitable for transparent, reproducible research and educational settings.

In contrast, CausalML[TM], developed by Uber, is more application-oriented, providing an all-inclusive approach for uplift modeling and heterogeneous treatment effect estimation. It supports several advanced algorithms including meta-learners (T-, S-, X-, R-learners), Uplift Random Forests, and implementations for multiple treatment groups. While it can be used with observational data, CausalML[TM] is particularly geared toward randomized experimental data, enabling fine-grained personalization and targeting strategies—especially in business contexts like marketing, customer engagement, and causal impact analysis.

### 4.1.2 Differences and Relative Merits

A key difference lies in their core philosophy and use case. PyWhy[TM] is designed for robustness and methodological rigor. It appeals to researchers and data scientists aiming to understand and validate causal assumptions in a structured way. By requiring users to explicitly define causal models, PyWhy[TM] promotes transparency and critical reasoning but may require more domain expertise.

CausalML[TM], on the other hand, focuses on scalability and ease of use, especially for industry applications. Its consistent API mirrors that of standard machine learning libraries, allowing practitioners to estimate treatment effects with minimal code. It's particularly strong in supporting uplift modeling, which is crucial for real-time decision-making systems. However, CausalML[TM]'s emphasis on performance over causal model specification can lead to overreliance on black-box models unless users remain cautious about confounding and selection bias, especially in observational settings.

### 4.1.3 Challenges and Limitations

Both tools face limitations reflective of broader challenges in causal inference. For PyWhy[TM], while it offers a robust pipeline, it may lack scalability for large-scale or high-dimensional data due to its limited support for newer machine learning methods like deep learning. Its usability can also be a barrier for non-specialists, particularly when defining causal graphs or conducting refutations.

CausalML[TM] excels in handling large-scale data and complex treatment setups, but its lack of built-in support for graphical models and formal assumption testing may make it vulnerable to biased inference if users fail to account for confounding variables correctly. Moreover, while powerful, uplift models can be data-hungry and sensitive to imbalance, particularly in cases with sparse treatment-response pairs.

## 4.2 Additional Tools Used

Pandas (McKinney & others, 2011) is a Python library with rich data structures and tools for working with structured datasets that are common in statistics, finance, mathematics, and social sciences and many other fields. The library provides comprehensive, intuitive routines for performing common manipulations and analyses on such datasets. It serves as a powerful complement to the existing Python scientific stack, while implementing and enhancing the data manipulation tools found in other statistical programming languages such as the R language.

Matplotlib (Hunter, 2007) is a Python two-dimensional graph library that produces quality publication layouts in a variety of print formats and interactive environments across all platforms. Matplotlib can be used in Python scripts, Python and IPython shells, the Jupyter notebook (Kluyver et al., 2016), web application servers and in GUI toolkits. Some of the graphical representations that can be created are histograms, power spectra, bar charts, diagrams error bars, scatter plots, etc.

Diagrams.net (JGraph, 2005) is an open source solution for creating sketches and diagrams. The tool can be used online with various storage platforms (Atlassian Confluence Cloud and Jira Cloud, applications of Google, GitHub and Microsoft applications) and offline as a standalone application. Unofficial integrations exist and are available for many other platforms and tools.

## 5 The Case Study

### 5.1 The Dataset

The IDEAL Household Energy Dataset (Pullinger et al., 2021) was used, which includes measurements from 255 homes, such as electricity consumption at a sampling period of 1 sec, pulse-level gas data, temperature, humidity and lighting data at a sampling period of 12 secs for each room and data temperature from the boiler tubes concerning central heating and hot water at a sampling period of 12 secs. 39 of the homes also include plug-level monitoring of selected electrical appliances, real power measurement of the mains and main sub-circuits, and more detailed temperature monitoring of equipment that uses gas and heat, including radiators and taps. The data set additionally includes demographics, values and self-reported energy awareness of residents, household income, as well as building and space characteristics and a record of energy-intensive appliances. Secondary data include weather conditions and level of urbanization.

The project was funded by the UK Engineering and Physical Sciences Research Council through two initiatives and received ethical approval from the University of Edinburgh. All equipment met CE certification and was professionally installed.

Data fields are documented in the IDEALdata.md file. The dataset is available in CSV format, organized by household, room, and sensor ID. A metadata structure links participants to homes, rooms, sensor boxes, and sensors.

### 5.2 Data Preprocessing

The first step in preprocessing was quantifying data availability to decide which homes were suitable for inclusion in model training. Homes with at least six months of data on core features (electricity/gas use, temperature, humidity, lighting) were selected. Ultimately, 212 homes were used for electricity and 208 for gas consumption analysis.

Next, various features were computed for each home, such as average consumption during cold (Sept–Mar) and warm periods (July–Sept, Mar–July), and over the entire year. Additional engineered features were also included based on their causal relevance to the research questions.

These included:
- Number of external windows, doors, and walls (distinguished by whether they could be opened),
- Average room area and height, and total house volume,
- Proportion of space used for drying clothes,
- Percentage of space with thermostatic radiator valves (TRVs),
- Average gas and electricity prices per kWh.

Categorical data was numerically encoded (e.g., one-hot encoding), and missing values were imputed using average values.

Final datasets were compiled into CSV files for different seasonal periods. Specific labels were generated for answering research questions, such as binary indicators of high electricity or gas consumption. Each home's average usage was compared to the overall average, with labels assigned based on whether it was above or below this benchmark.

Additional questions were addressed similarly, including income level, number of windows, or number of electric heaters. Labels like high_income or many_electric_heaters reflect the presence of corresponding features.

Finally, Directed Acyclic Graphs (DAGs) were created using Diagrams.net to represent causal relationships for each question. These were then translated for use with the PyWhy[TM] tool. Unobserved confounding variables were marked as "U" where necessary. Outputs from both tools were consolidated into tables and JSON files for visualization and comparison.

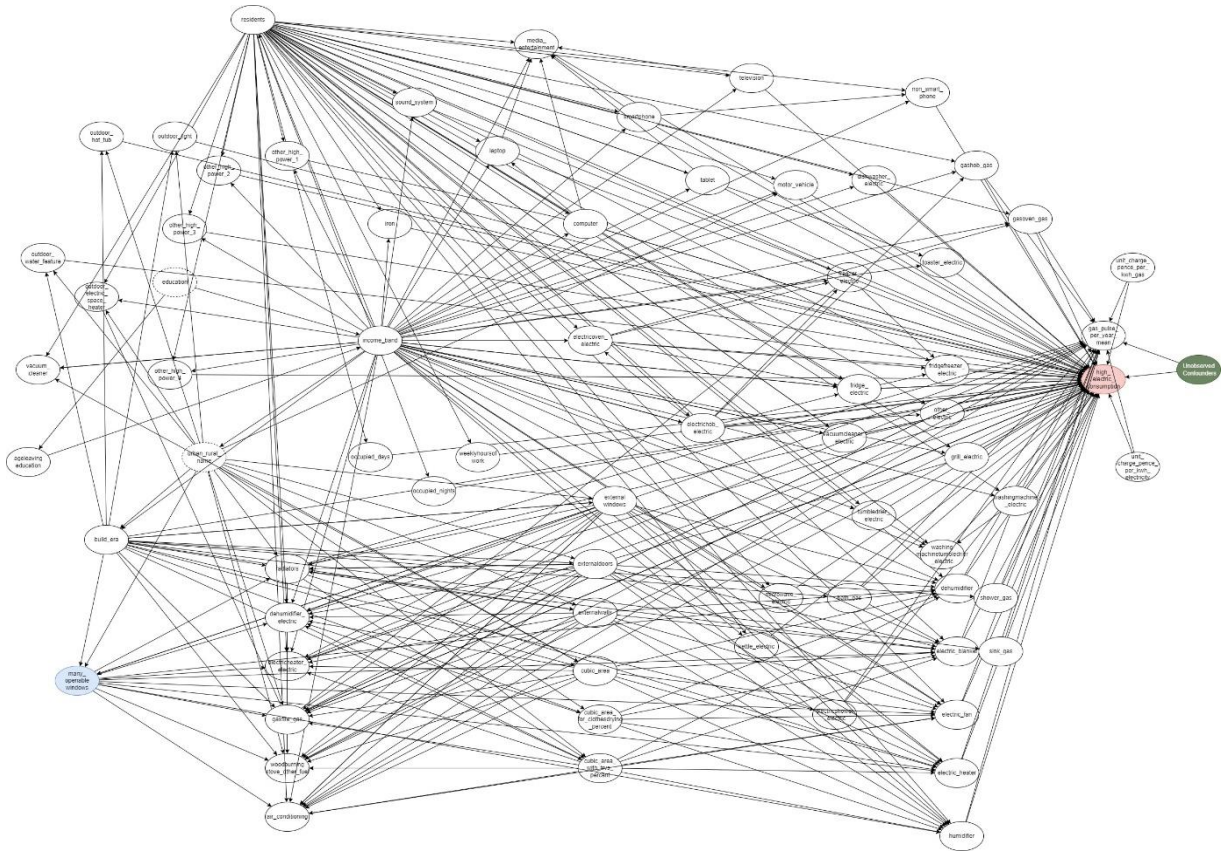

*Fig. 1 The visualised DAG fed into the PyWhy[TM] tool to answer question 4 (electricity consumption). The treatment variable is marked in red, the outcome variable in blue, and the unobserved confounders in green.*

### 5.3 Queries

The 18 queries formulated are presented in Table 1.

*Table 1 CML queries and their usage context (architectural design changes - ADC, engineering interventions - EI, alternative uses of rooms- AUR, social intervention - SI, environmental conditions – EC, avg: average)*

|   | Output | Input | Usage |
|---|---|---|---|
| 1 |  | high source of household income | SI |
| 2 |  | high level of education among the occupants | SI |
| 3 |  | large number of windows | ADC |
| 4 | Change in average consumption | large number of openable windows | EI |
| 5 |  | large number of exterior doors | ADC |
| 6 |  | a large number of exterior walls | ADC |
| 7 |  | large volume of the rooms | ADC |

| | Output | Input | Usage |
|---|---|---|---|
| 8 | | large area of the rooms | ADC |
| 9 | | high walls of the rooms | ADC |
| 10 | | a large space (volume) for drying clothes | AU |
| 11 | | large space with built-in thermostatic valves in the radiators | AU & EI |
| 12 | | many tenants | AU |
| 13 | | house located in an urban area | EC |
| 14 | | change of time to winter time | EC |
| 15 | Change in avg consumption in winter | many electric heaters in the house | EI |
| 16 | Change in avg temperature in winter | many electric heaters in the house | EI |
| 17 | Change in avg consumption in summer | existence of electric fans or air conditioners | EI |
| 18 | Change in avg temperature in summer | existence of electric fans or air conditioners | EI |

# 6  Results and Discussion

## 6.1  Results

The study utilizes meta-learners from the CausalML[TM] library (such as T-Learner, S-Learner, X-Learner, and R-Learner) and sensitivity analysis techniques from the PyWhy[TM] library. Both libraries incorporate diverse model architectures, notably Linear Regression and XGBoost (Extreme Gradient Boosting). Due to convergence issues in CausalML[TM], a cap of 100 iterations was imposed; however, increasing this threshold yielded negligible changes in estimates.

A typical result for one of the 18 queries, namely for query No 4, is presented in Figure 1. The rest 17 queries yield analogous results that can be seen in "Appendix" section. Question 4 analyzed whether households with a high number of openable windows exhibit different patterns in electricity usage. Homes exceeding the dataset's average of 5 openable windows were classified as having a high number of openable windows.

Analysis across multiple thresholds for classifying a home as "window-rich" showed a consistent trend: consumption increased when the number of windows exceeded five. Below that threshold, the probability decreased. This pattern suggests thermal inefficiency from excessive window exposure, offset by increased HVAC usage.

The study highlights the influence of confounders such as occupant environmental behavior and window size, advocating for richer datasets to refine estimates.

Findings across methodologies were as follows:

- Propensity Score Matching and Linear Regression via PyWhy[TM] estimated a 24% increase in consumption.
- Inverse Propensity Weighting and Propensity Score Stratification in PyWhy[TM] yielded higher estimates (43%).
- CausalML[TM] meta-learners produced diverse results: the XGBoost-based T-Learner aligned closely with PyWhy[TM] (23–24%), while other learners (S, X, R) diverged significantly (3–15%).

The MLP model was excluded from plots due to large deviations across all queries.

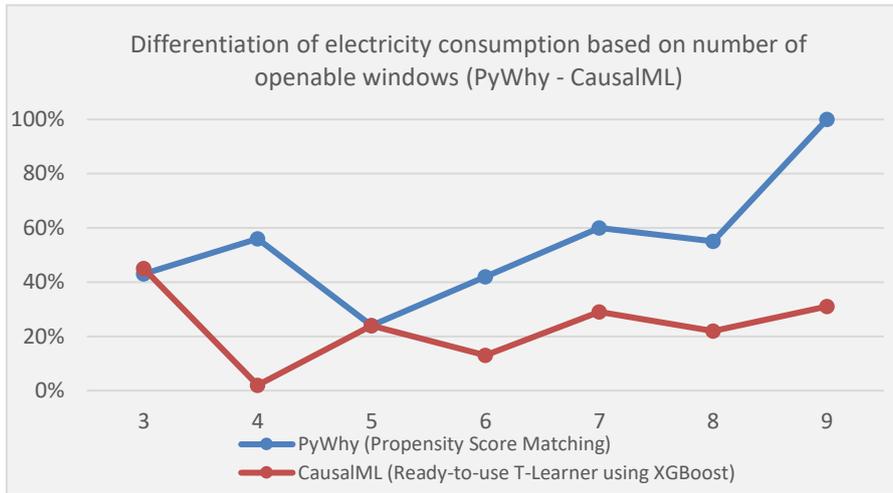

Fig. 2 Comparative graph of PyWhy[TM] and CausalML[TM] results for different number of openable windows.

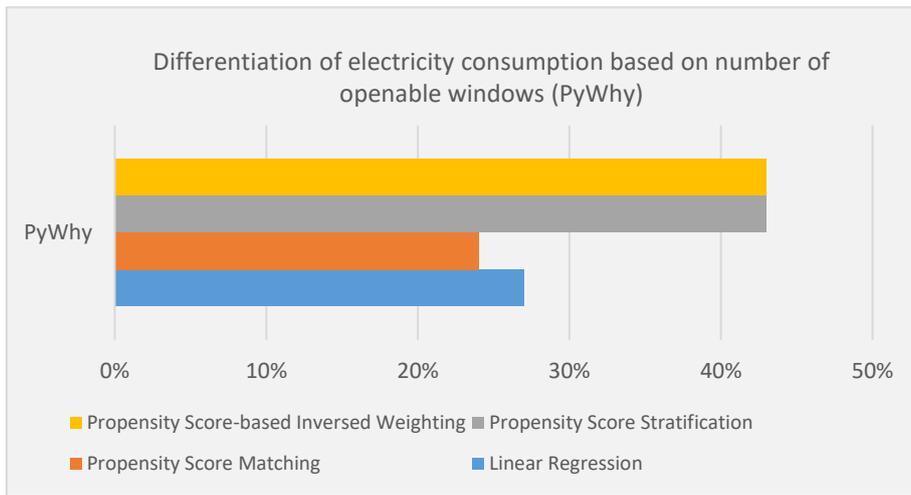

Fig. 3 Results of query 4 regarding the PyWhy[TM] tool.

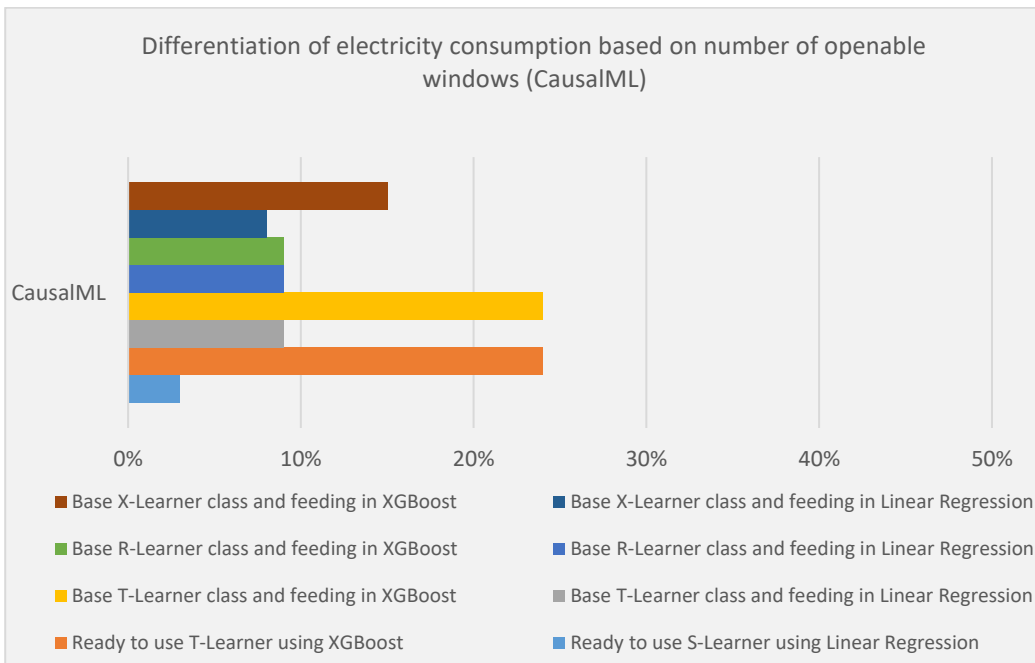

Fig. 4 Results of query 4 regarding the CausalML[TM] tool.

Since true counterfactuals are unobservable outside randomized trials, validation hinges on internal consistency, notably via Sensitivity Analysis. This evaluates the robustness of treatment effect estimates (Average Treatment Effect on the Treated - ATT) under hypothetical unobserved confounding. AUUC is also employed to assess performance, particularly regarding how well uplift models capture treatment heterogeneity.

### 6.1.1 Sensitivity Analysis

Sensitivity checks in PyWhy™ indicated that:

- Replacing the dataset with a random subset or adding a random common cause marginally changed ATT estimates (~1% change), affirming causal assumptions.
- Introducing unobserved confounders caused an 11% variation.
- Replacing the treatment variable with a random independent one resulted in a 27% drop (towards zero), indicating strong causal relevance of the original treatment.

Associated p-values further validated these shifts, with the placebo test showing high p-values (~2) and other transformations yielding values from 0.13 to 0.9.

CausalML™ produced analogous findings, except for tests involving unobserved confounders, which are not supported by that library. Overall, both tools corroborated the validity of the causal claims.

### 6.1.2 Performance on Synthetic Data

To overcome the absence of known ground truth in the real dataset, a synthetic dataset of 10,000 samples was generated (repeated 10 times), with training/validation split (80/20). The function simulate_nuisance_and_easy_treatment was used to simulate data with known causal structures.

Performance assessments revealed that:

- XGBoost-based meta-learners (X, T, R) consistently outperformed others in estimating ATT.
- Linear Regression-based learners underperformed, with S-Learner performing worst.
- MSE was lower for training and validation sets using linear models but less accurate regarding true ATT values.
- KL Divergence analyses indicated S-Learner with Linear Regression had significantly higher distributional divergence (~12×) from the true data, marking it as the least reliable.

These results underscore XGBoost's superior capacity to model complex causal relationships under non-linear and high-dimensional contexts.

### 6.1.3 Uplift Curve Validation

Uplift modeling quantifies how treatment influences individual outcomes. AUUC provides a summary performance metric by measuring the difference in responses between treated and control groups.

- XGBoost-based learners again exhibited the highest AUUC values for both training and validation sets.
- Linear Regression models showed poor uplift performance, especially the S-Learner, which failed to approach the ideal prediction line (y = x).
- Visual comparisons of predicted vs. actual outcomes reaffirmed that only the XGBoost learners closely approximated the ideal causal estimator.

## 6.2 Assumptions and Limitations

Several assumptions underlie the findings:

- Hourly averages were used to approximate monthly values due to limited computational resources. Missing data might bias estimates.
- Sensor placement inconsistencies (e.g., height of temperature sensors) were not accounted for, possibly distorting real indoor temperature readings.
- Zero gas consumption over long periods may stem from lack of use or sensor/system faults, further complicating inference.

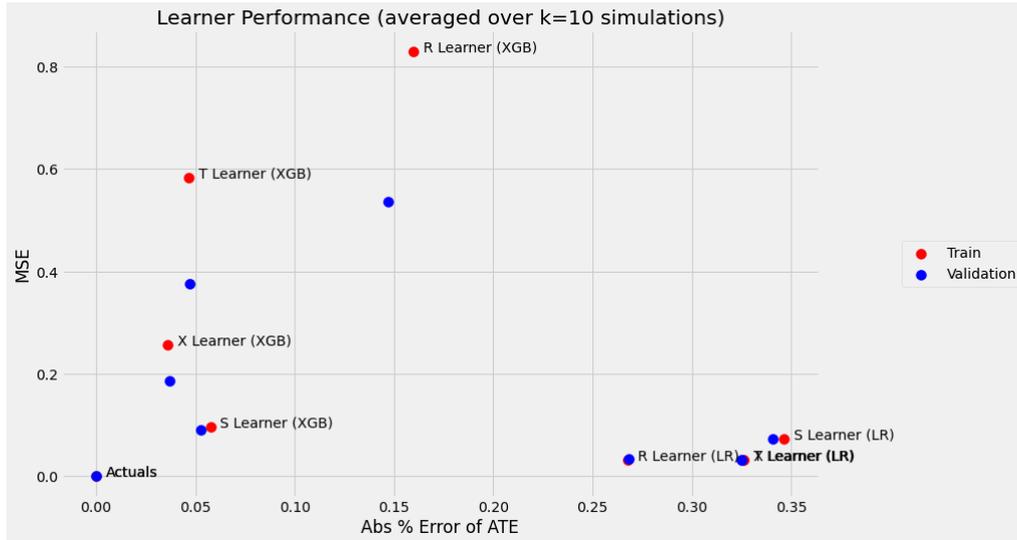

*Fig. 5 Scatter plot of meta-learner performance.*

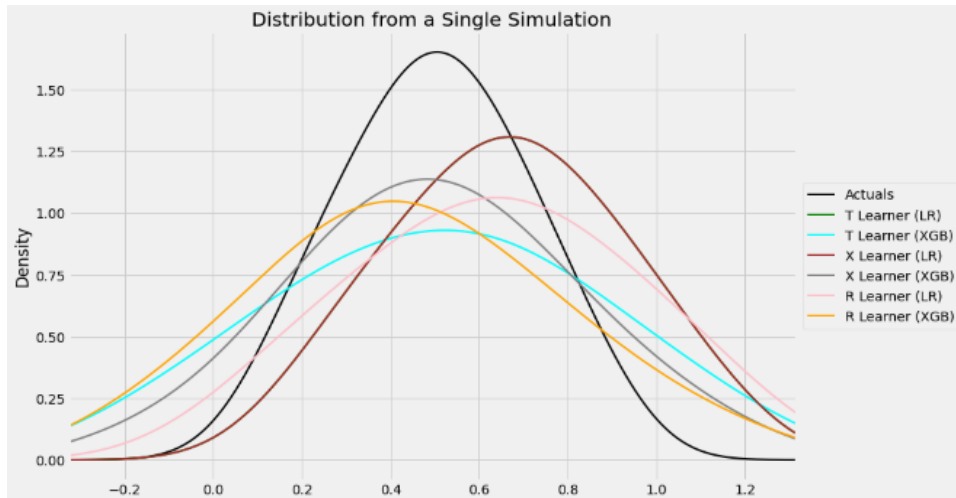

*Fig. 6 Probability distributions of validation datasets.*

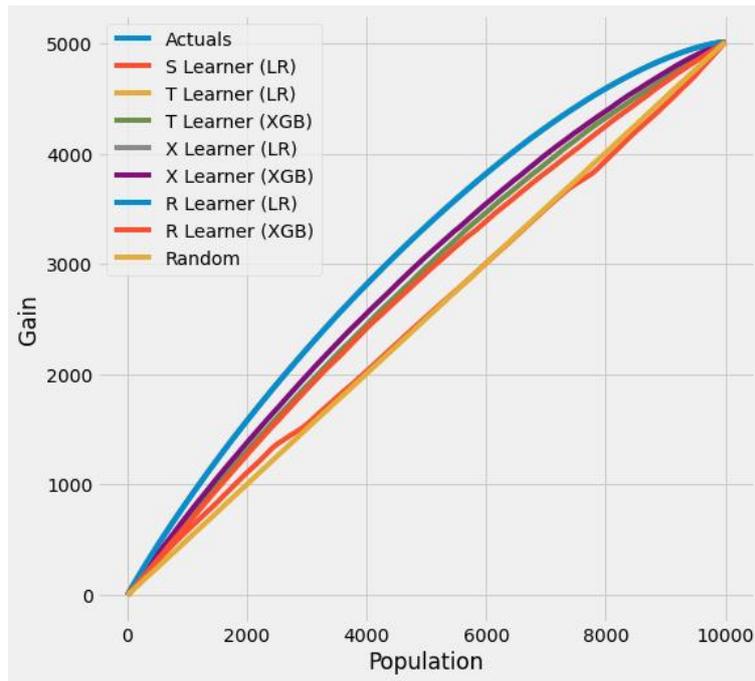

Fig. 7 Uplift curves for validation datasets.

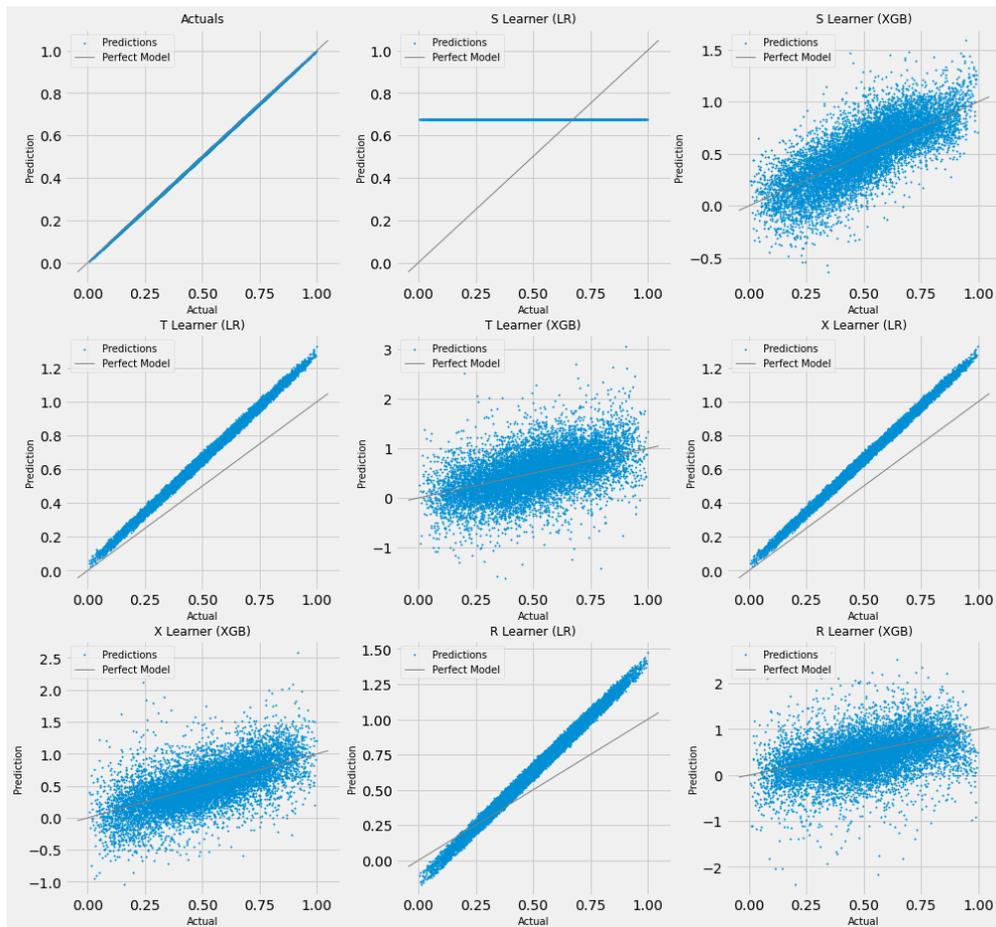

Fig. 8 Comparison between meta-learner predictions and a perfect model for validation datasets.

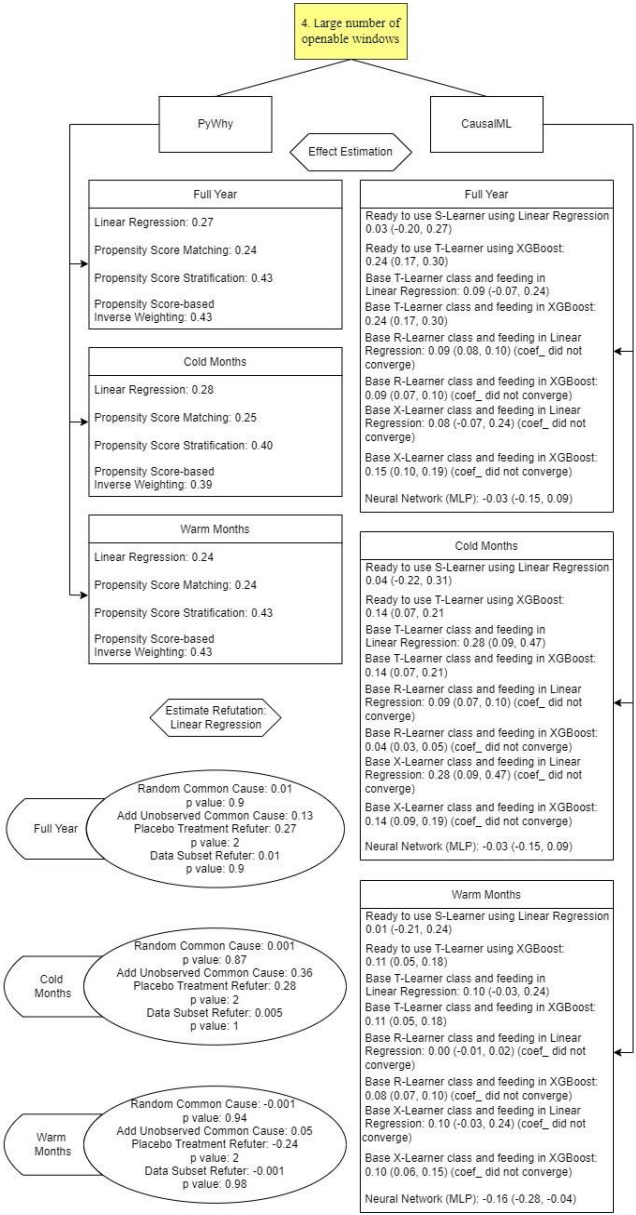
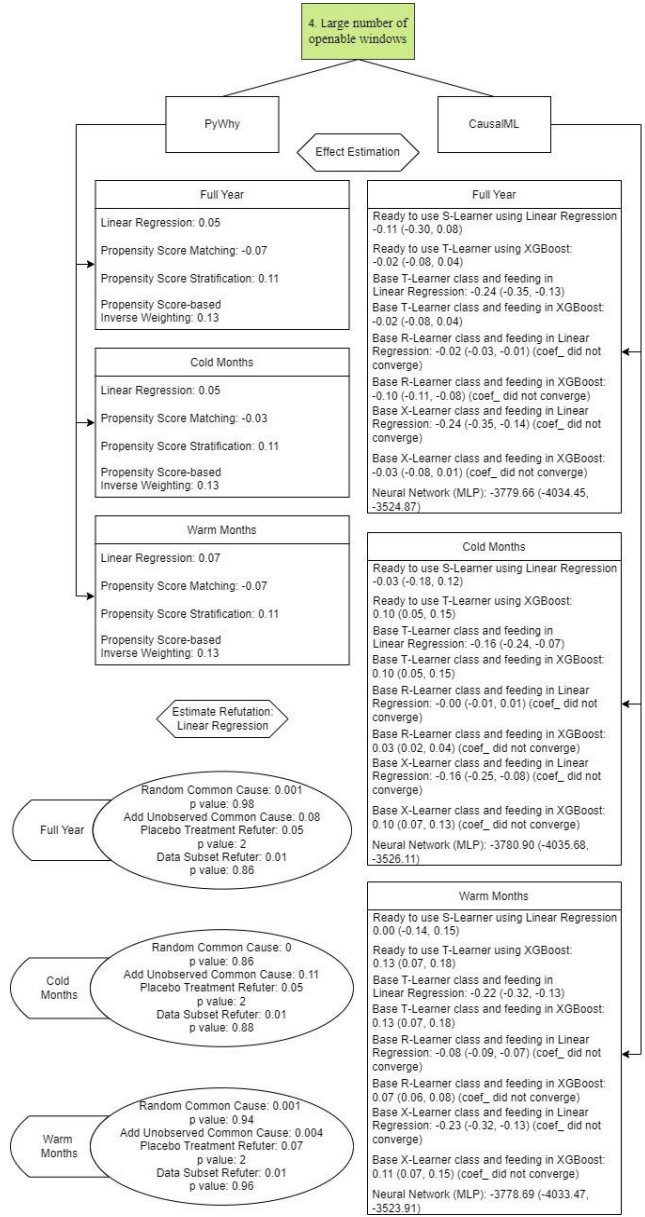

Fig. 9 Analytic results of query 4 (electricity usage)  Fig. 10 Analytic results of query 4 (gas usage)

## 7 Conclusions and Future Work

The slight divergence of the results of the two tools points to the field knowledge required to design the structural causal model fed into PyWhy[TM], this not necessarily leading to a better result than CausalML[TM].

Additionally, internal validation methods are able to assess the causality assumptions we considered, but not the treatment effect which is rendered impossible by having the true value only for experimental data.

For both PyWhy[TM] and CausalML[TM] it is important to study the estimated ICU of each algorithm, in order to choose the domain-specific optimum. It is also important to evaluate the assumptions of the causal relationship based on the pre-existing domain knowledge, via the internal validation methods provided.

The tools used in this work can be improved, mainly in terms of a friendlier user experience. More specifically, it would be useful to include in PyWhy[TM] a user interface for creating the structural causal model and automatically translating it into code that the tool processes, thus avoiding its time-consuming visualization with a third-party program and its manual conversion into recognizable code from PyWhy$^{TM}$.

Some suggestions for future work on data preprocessing are as follows:

- Using a larger data set to train the causal model with the aim of deriving more reliable results. For example, of the 212 houses in the used data set, 43 were rejected due to incomplete data.
- Collecting data from different locations, in order to diversify the data and draw more individualized conclusions.
- Calculating HCV outside the ICU to reveal heterogeneity between population subgroups, aiming at more individualized outcomes and, by extension, enforcing more targeted treatment by subgroup.

In addition, suggestions for future work regarding outcome evaluation are as follows:

- Application of the results produced in real conditions (in this case, e.g., changing the architecture of new houses, the distribution of electrical appliances per house), aiming at their practical evaluation.
- Trying additional CML tools such as the CausalLib (Shimoni et al., 2019) and Causalinference (Wong, 2014) libraries of the Python language.

# Appendix

In this section the results of the remaining 17 queries are provided, as categorized by table 1. Results of queries regarding electricity consumption are distinguished by yellow heading, natural gas consumption by green heading and temperature by blue heading.

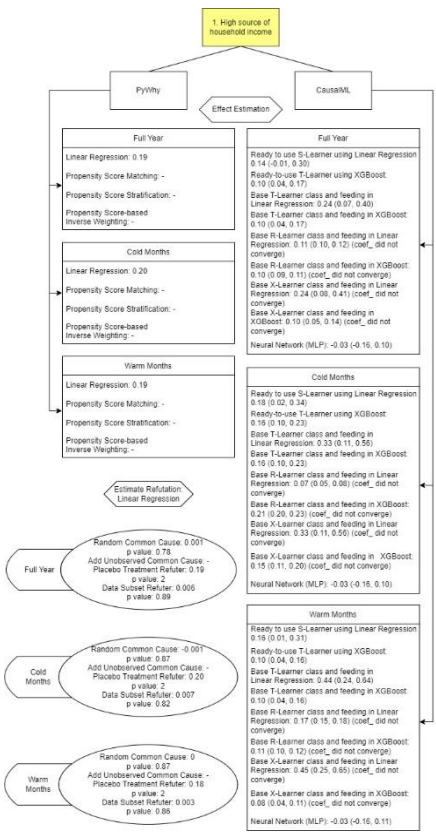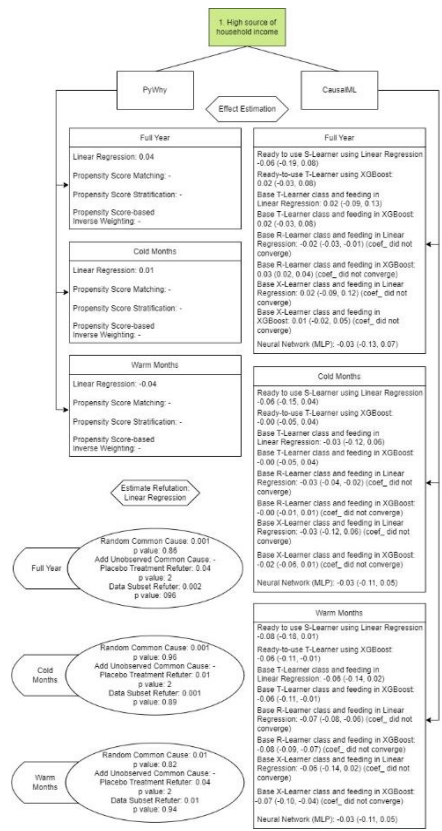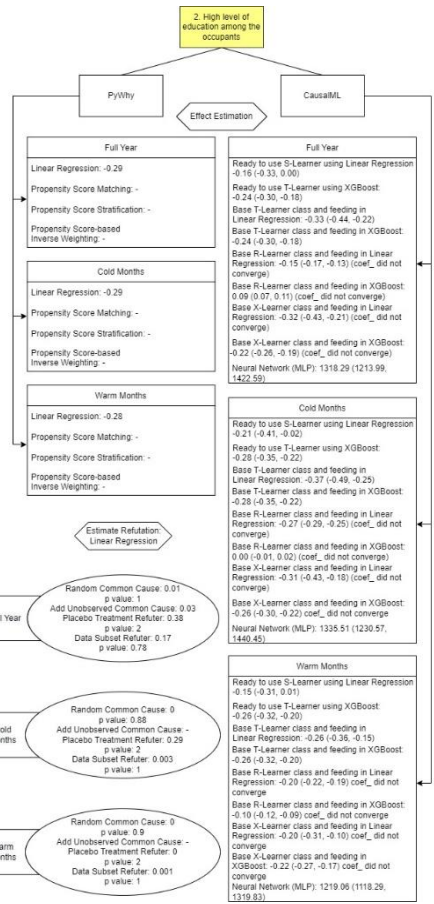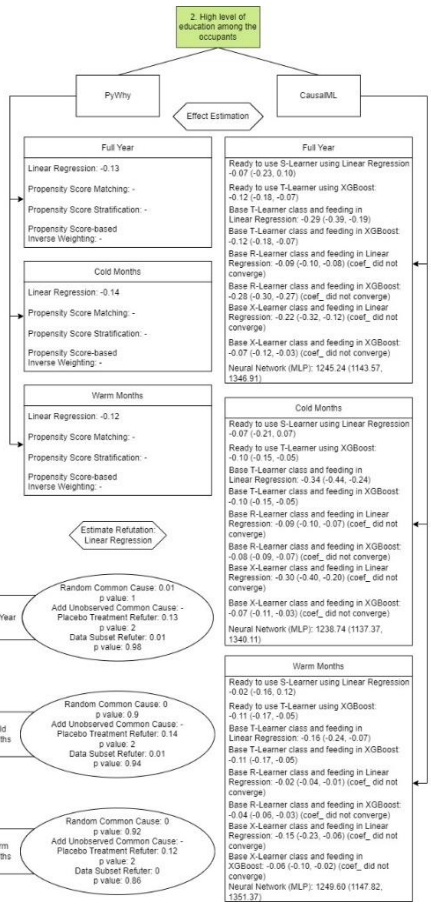

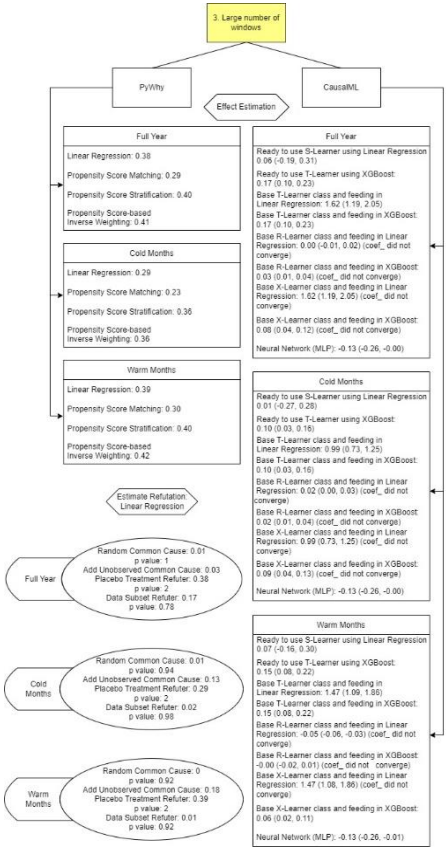
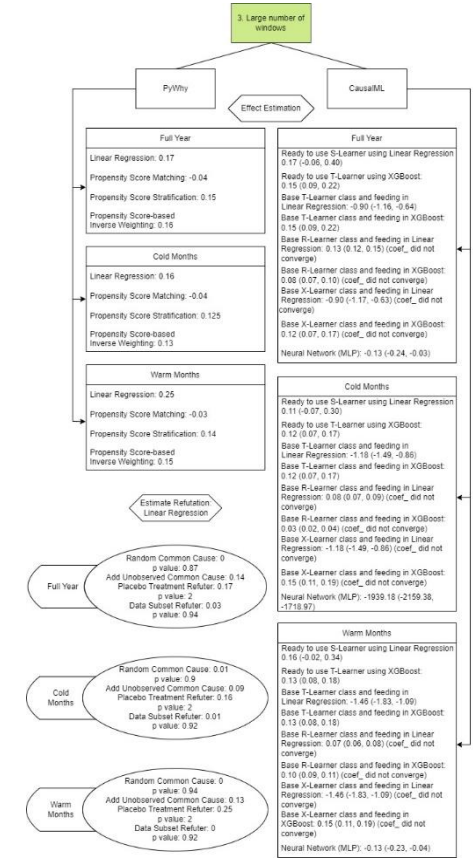
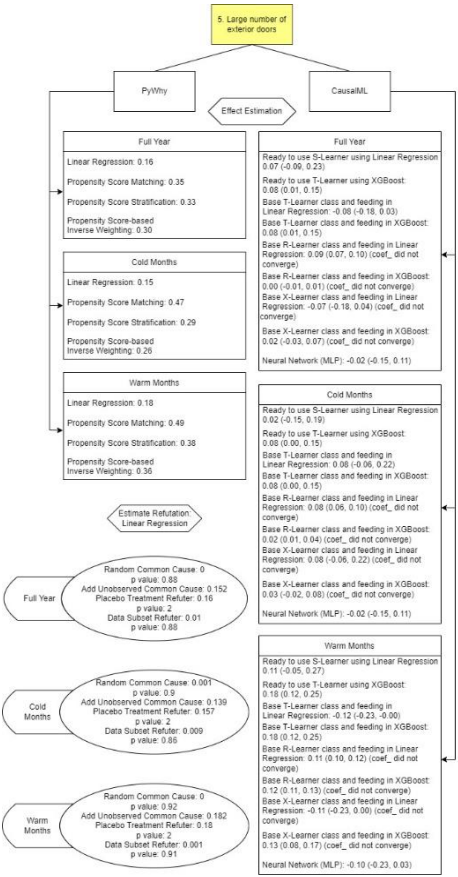
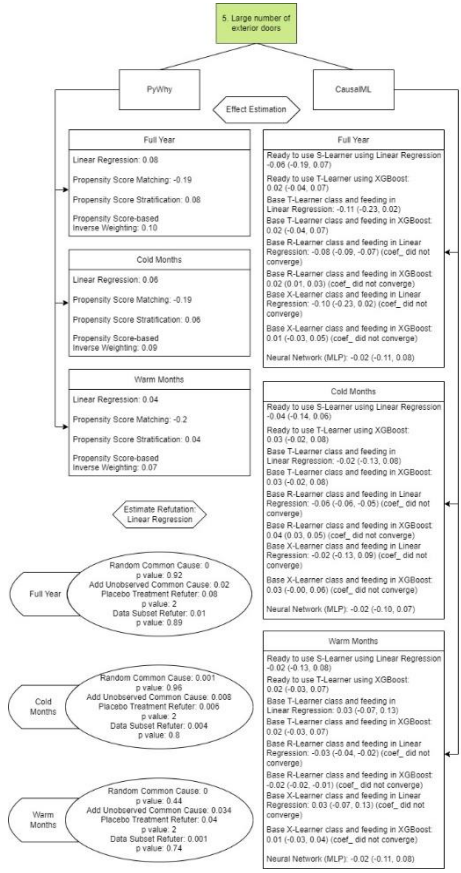

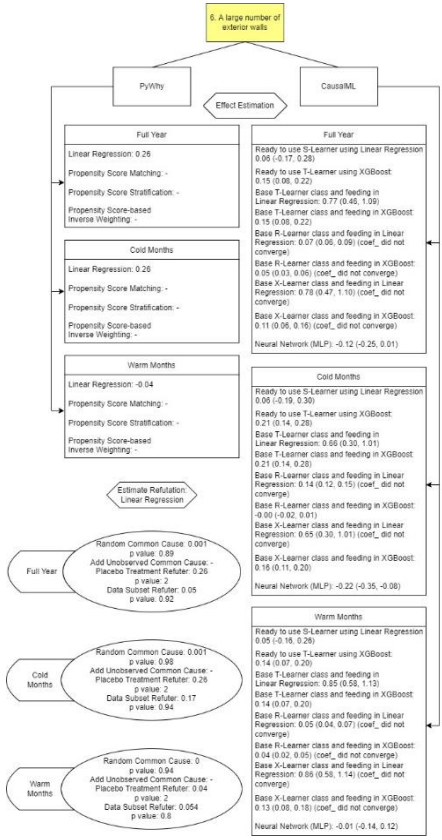
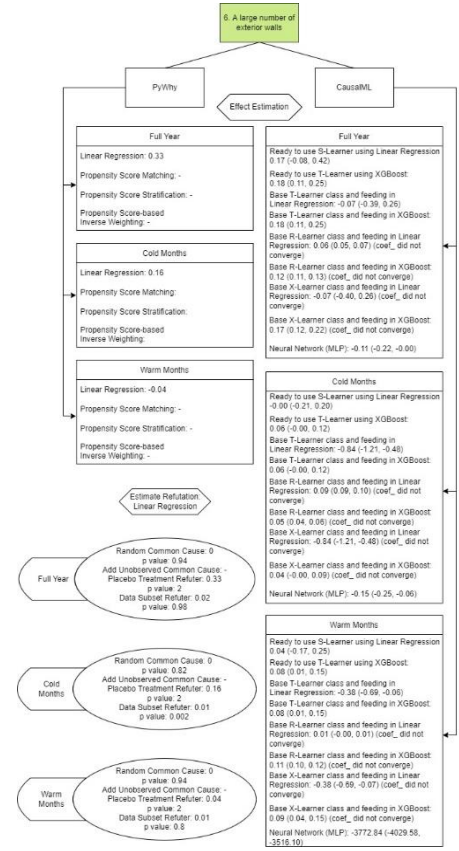
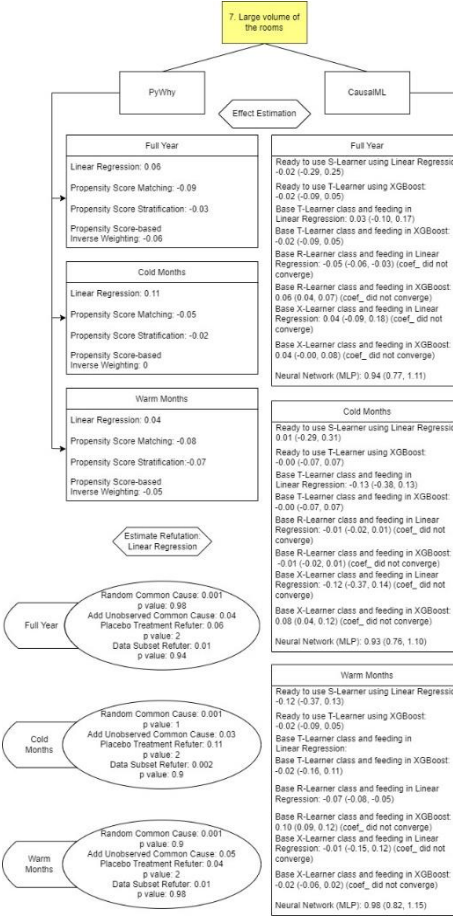
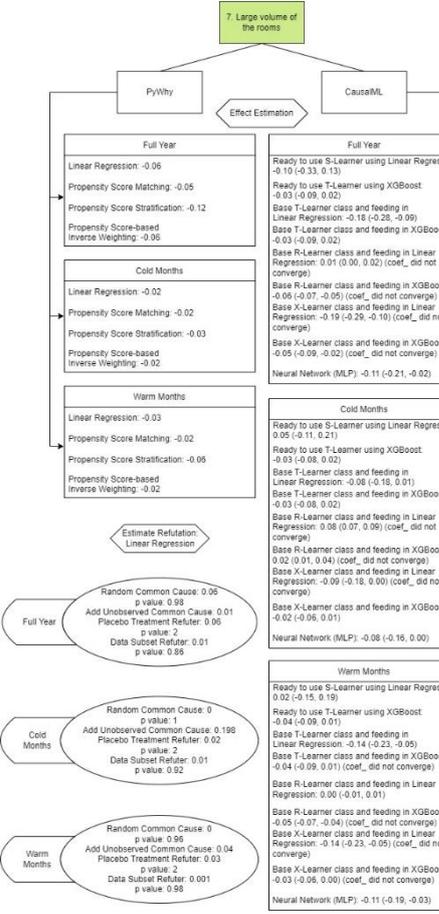

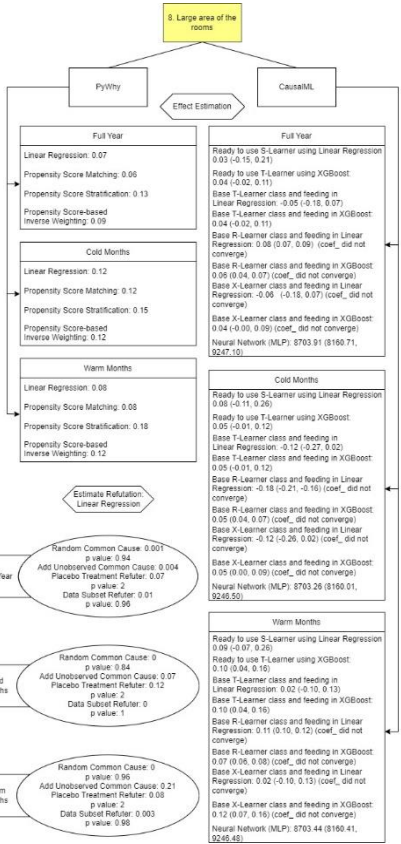
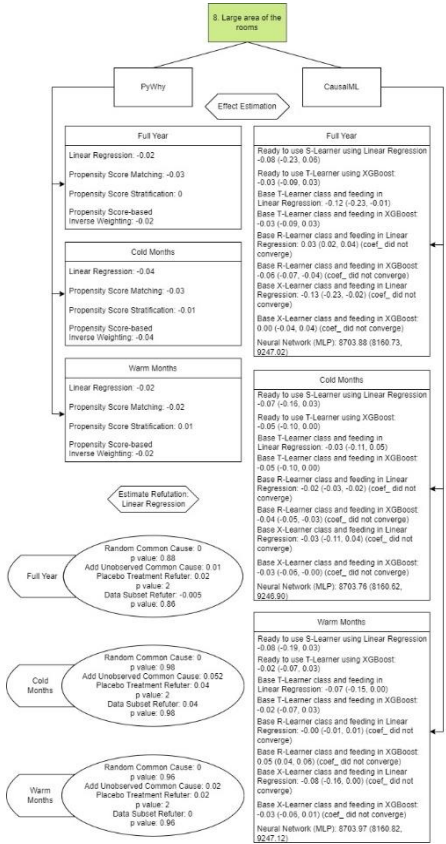
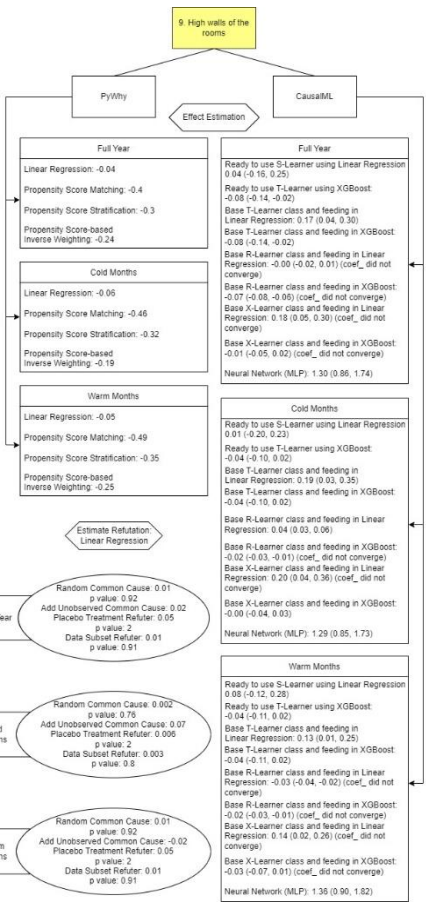
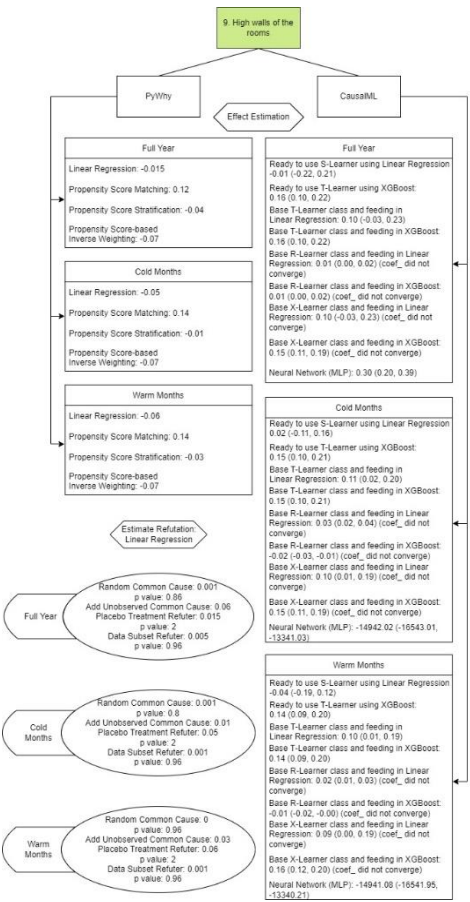

## 10. A large space (volume) for drying clothes

### PyWhy — Effect Estimation — CausalML

**Full Year**
- Linear Regression: No relevant identified estimand
- Propensity Score Matching: No relevant identified estimand
- Propensity Score Stratification: No relevant identified estimand
- Propensity Score-based Inverse Weighting: No relevant identified estimand

**Cold Months**
- Linear Regression: No relevant identified estimand
- Propensity Score Matching: No relevant identified estimand
- Propensity Score Stratification: No relevant identified estimand
- Propensity Score-based Inverse Weighting: No relevant identified estimand

**Warm Months**
- Linear Regression: No relevant identified estimand
- Propensity Score Matching: No relevant identified estimand
- Propensity Score Stratification: No relevant identified estimand
- Propensity Score-based Inverse Weighting: No relevant identified estimand

**CausalML — Full Year**
- Ready to use S-Learner using Linear Regression: -0.15 (-0.29, -0.02)
- Ready to use T-Learner using XGBoost: -0.08 (-0.14, -0.01)
- Base T-Learner class and feeding in Linear Regression: -0.12 (-0.26, 0.03)
- Base T-Learner class and feeding in XGBoost: -0.08 (-0.14, -0.01)
- Base R-Learner class and feeding in Linear Regression: -0.17 (-0.18, -0.16) (coef_ did not converge)
- Base R-Learner class and feeding in XGBoost: -0.08 (-0.09, -0.07) (coef_ did not converge)
- Base X-Learner class and feeding in Linear Regression: -0.12 (-0.27, 0.03) (coef_ did not converge)
- Base X-Learner class and feeding in XGBoost: -0.06 (-0.09, -0.02) (coef_ did not converge)
- Neural Network (MLP): 0.35 (0.22, 0.48)

**CausalML — Cold Months**
- Ready to use S-Learner using Linear Regression: -0.15 (-0.30, -0.01)
- Ready to use T-Learner using XGBoost: -0.11 (-0.18, -0.05)
- Base T-Learner class and feeding in Linear Regression: -0.19 (-0.36, -0.02)
- Base T-Learner class and feeding in XGBoost: -0.11 (-0.18, -0.05)
- Base R-Learner class and feeding in Linear Regression: -0.21 (-0.22, -0.19) (coef_ did not converge)
- Base R-Learner class and feeding in XGBoost: -0.16 (-0.17, -0.15) (coef_ did not converge)
- Base X-Learner class and feeding in Linear Regression: -0.20 (-0.37, -0.02) (coef_ did not converge)
- Base X-Learner class and feeding in XGBoost: -0.08 (-0.12, -0.05) (coef_ did not converge)
- Neural Network (MLP): 0.35 (0.22, 0.47)

**CausalML — Warm Months**
- Ready to use S-Learner using Linear Regression: -0.20 (-0.33, -0.07)
- Ready to use T-Learner using XGBoost: -0.10 (-0.15, -0.04)
- Base T-Learner class and feeding in Linear Regression: -0.16 (-0.30, -0.03)
- Base T-Learner class and feeding in XGBoost: -0.10 (-0.15, -0.04)
- Base R-Learner class and feeding in Linear Regression: -0.20 (-0.21, -0.18) (coef_ did not converge)
- Base R-Learner class and feeding in XGBoost: -0.17 (-0.18, -0.16) (coef_ did not converge)
- Base X-Learner class and feeding in Linear Regression: -0.16 (-0.30, -0.03) (coef_ did not converge)
- Base X-Learner class and feeding in XGBoost: -0.09 (-0.12, -0.06) (coef_ did not converge)
- Neural Network (MLP): 0.35 (0.22, 0.48)

### Estimate Refutation: Linear Regression

**Full Year**
- Random Common Cause: - p value: -
- Add Unobserved Common Cause: -
- Placebo Treatment Refuter: - p value: -
- Data Subset Refuter: - p value: -

**Cold Months**
- Random Common Cause: - p value: -
- Add Unobserved Common Cause: -
- Placebo Treatment Refuter: - p value: -
- Data Subset Refuter: - p value: -

**Warm Months**
- Random Common Cause: - p value: -
- Add Unobserved Common Cause: -
- Placebo Treatment Refuter: - p value: -
- Data Subset Refuter: - p value: -

---

## 10. A large space (volume) for drying clothes

### PyWhy — Effect Estimation — CausalML

**Full Year**
- Linear Regression: No relevant identified estimand
- Propensity Score Matching: No relevant identified estimand
- Propensity Score Stratification: No relevant identified estimand
- Propensity Score-based Inverse Weighting: No relevant identified estimand

**Cold Months**
- Linear Regression: No relevant identified estimand
- Propensity Score Matching: No relevant identified estimand
- Propensity Score Stratification: No relevant identified estimand
- Propensity Score-based Inverse Weighting: No relevant identified estimand

**Warm Months**
- Linear Regression: No relevant identified estimand
- Propensity Score Matching: No relevant identified estimand
- Propensity Score Stratification: No relevant identified estimand
- Propensity Score-based Inverse Weighting: No relevant identified estimand

**CausalML — Full Year**
- Ready to use S-Learner using Linear Regression: 0.11 (-0.01, 0.24)
- Ready to use T-Learner using XGBoost: 0.09 (0.03, 0.14)
- Base T-Learner class and feeding in Linear Regression: 0.07 (-0.06, 0.21)
- Base T-Learner class and feeding in XGBoost: 0.09 (0.03, 0.14)
- Base R-Learner class and feeding in Linear Regression: 0.01 (0.00, 0.02) (coef_ did not converge)
- Base R-Learner class and feeding in XGBoost: 0.11 (0.10, 0.12) (coef_ did not converge)
- Base X-Learner class and feeding in Linear Regression: 0.08 (-0.05, 0.21) (coef_ did not converge)
- Base X-Learner class and feeding in XGBoost: 0.09 (0.06, 0.13) (coef_ did not converge)
- Neural Network (MLP): 1942.82 (1691.82, 2193.82)

**CausalML — Cold Months**
- Ready to use S-Learner using Linear Regression: 0.09 (-0.02, 0.20)
- Ready to use T-Learner using XGBoost: 0.08 (0.03, 0.13)
- Base T-Learner class and feeding in Linear Regression: 0.04 (-0.05, 0.14)
- Base T-Learner class and feeding in XGBoost: 0.08 (0.03, 0.13)
- Base R-Learner class and feeding in Linear Regression: 0.07 (0.06, 0.08) (coef_ did not converge)
- Base R-Learner class and feeding in XGBoost: 0.10 (0.09, 0.11) (coef_ did not converge)
- Base X-Learner class and feeding in Linear Regression: 0.05 (-0.05, 0.14) (coef_ did not converge)
- Base X-Learner class and feeding in XGBoost: 0.08 (0.05, 0.11) (coef_ did not converge)
- Neural Network (MLP): 1945.10 (1693.95, 2196.26)

**CausalML — Warm Months**
- Ready to use S-Learner using Linear Regression: 0.09 (-0.02, 0.19)
- Ready to use T-Learner using XGBoost: 0.01 (-0.04, 0.06)
- Base T-Learner class and feeding in Linear Regression: 0.04 (-0.05, 0.13)
- Base T-Learner class and feeding in XGBoost: 0.01 (-0.04, 0.06)
- Base R-Learner class and feeding in Linear Regression: 0.04 (0.03, 0.05) (coef_ did not converge)
- Base R-Learner class and feeding in XGBoost: -0.06 (-0.07, -0.05) (coef_ did not converge)
- Base X-Learner class and feeding in Linear Regression: 0.04 (-0.05, 0.13) (coef_ did not converge)
- Base X-Learner class and feeding in XGBoost: 0.01 (-0.02, 0.05) (coef_ did not converge)
- Neural Network (MLP): 1941.10 (1690.28, 2191.91)

### Estimate Refutation: Linear Regression

**Full Year**
- Random Common Cause: - p value: -
- Add Unobserved Common Cause: -
- Placebo Treatment Refuter: - p value: -
- Data Subset Refuter: - p value: -

**Cold Months**
- Random Common Cause: - p value: -
- Add Unobserved Common Cause: -
- Placebo Treatment Refuter: - p value: -
- Data Subset Refuter: - p value: -

**Warm Months**
- Random Common Cause: - p value: -
- Add Unobserved Common Cause: -
- Placebo Treatment Refuter: - p value: -
- Data Subset Refuter: - p value: -

---

## 11. Large space with built-in thermostatic valves in the radiators

### PyWhy — Effect Estimation — CausalML

**Full Year**
- Linear Regression: No relevant identified estimand
- Propensity Score Matching: No relevant identified estimand
- Propensity Score Stratification: No relevant identified estimand
- Propensity Score-based Inverse Weighting: No relevant identified estimand

**Cold Months**
- Linear Regression: No relevant identified estimand
- Propensity Score Matching: No relevant identified estimand
- Propensity Score Stratification: No relevant identified estimand
- Propensity Score-based Inverse Weighting: No relevant identified estimand

**Warm Months**
- Linear Regression: No relevant identified estimand
- Propensity Score Matching: No relevant identified estimand
- Propensity Score Stratification: No relevant identified estimand
- Propensity Score-based Inverse Weighting: No relevant identified estimand

**CausalML — Full Year**
- Ready to use S-Learner using Linear Regression: 0.07 (-0.06, 0.20)
- Ready to use T-Learner using XGBoost: -0.05 (-0.11, 0.01)
- Base T-Learner class and feeding in Linear Regression: -0.04 (-0.19, 0.11)
- Base T-Learner class and feeding in XGBoost: -0.05 (-0.11, 0.01)
- Base R-Learner class and feeding in Linear Regression: 0.03 (0.01, 0.04) (coef_ did not converge)
- Base R-Learner class and feeding in XGBoost: -0.05 (-0.06, -0.04) (coef_ did not converge)
- Base X-Learner class and feeding in Linear Regression: -0.04 (-0.19, 0.11) (coef_ did not converge)
- Base X-Learner class and feeding in XGBoost: -0.05 (-0.08, -0.01) (coef_ did not converge)
- Neural Network (MLP): 0.18 (0.04, 0.31)

**CausalML — Cold Months**
- Ready to use S-Learner using Linear Regression: 0.08 (-0.07, 0.23)
- Ready to use T-Learner using XGBoost: -0.00 (-0.07, 0.07)
- Base T-Learner class and feeding in Linear Regression: -0.11 (-0.31, 0.08)
- Base T-Learner class and feeding in XGBoost: -0.00 (-0.07, 0.07)
- Base R-Learner class and feeding in Linear Regression: -0.12 (-0.15, -0.09) (coef_ did not converge)
- Base R-Learner class and feeding in XGBoost: 0.01 (-0.01, 0.02) (coef_ did not converge)
- Base X-Learner class and feeding in Linear Regression: -0.11 (-0.31, 0.09) (coef_ did not converge)
- Base X-Learner class and feeding in XGBoost: -0.01 (-0.05, 0.04) (coef_ did not converge)
- Neural Network (MLP): 0.18 (0.05, 0.32)

**CausalML — Warm Months**
- Ready to use S-Learner using Linear Regression: 0.08 (-0.05, 0.21)
- Ready to use T-Learner using XGBoost: -0.00 (-0.07, 0.06)
- Base T-Learner class and feeding in Linear Regression: -0.09 (-0.30, 0.13)
- Base T-Learner class and feeding in XGBoost: -0.00 (-0.07, 0.06)
- Base R-Learner class and feeding in Linear Regression: 0.03 (0.02, 0.04) (coef_ did not converge)
- Base R-Learner class and feeding in XGBoost: 0.02 (0.00, 0.03) (coef_ did not converge)
- Base X-Learner class and feeding in Linear Regression: -0.09 (-0.30, 0.13) (coef_ did not converge)
- Base X-Learner class and feeding in XGBoost: -0.01 (-0.05, 0.02) (coef_ did not converge)
- Neural Network (MLP): 0.02 (-0.12, 0.15)

### Estimate Refutation: Linear Regression

**Full Year**
- Random Common Cause: - p value: -
- Add Unobserved Common Cause: -
- Placebo Treatment Refuter: - p value: -
- Data Subset Refuter: - p value: -

**Cold Months**
- Random Common Cause: - p value: -
- Add Unobserved Common Cause: -
- Placebo Treatment Refuter: - p value: -
- Data Subset Refuter: - p value: -

**Warm Months**
- Random Common Cause: - p value: -
- Add Unobserved Common Cause: -
- Placebo Treatment Refuter: - p value: -
- Data Subset Refuter: - p value: -

---

## 11. Large space with built-in thermostatic valves in the radiators

### PyWhy — Effect Estimation — CausalML

**Full Year**
- Linear Regression: No relevant identified estimand
- Propensity Score Matching: No relevant identified estimand
- Propensity Score Stratification: No relevant identified estimand
- Propensity Score-based Inverse Weighting: No relevant identified estimand

**Cold Months**
- Linear Regression: No relevant identified estimand
- Propensity Score Matching: No relevant identified estimand
- Propensity Score Stratification: No relevant identified estimand
- Propensity Score-based Inverse Weighting: No relevant identified estimand

**Warm Months**
- Linear Regression: No relevant identified estimand
- Propensity Score Matching: No relevant identified estimand
- Propensity Score Stratification: No relevant identified estimand
- Propensity Score-based Inverse Weighting: No relevant identified estimand

**CausalML — Full Year**
- Ready to use S-Learner using Linear Regression: -0.10 (-0.24, 0.04)
- Ready to use T-Learner using XGBoost: -0.06 (-0.12, -0.01)
- Base T-Learner class and feeding in Linear Regression: -0.25 (-0.45, -0.06)
- Base T-Learner class and feeding in XGBoost: -0.06 (-0.12, -0.01)
- Base R-Learner class and feeding in Linear Regression: -0.02 (-0.03, -0.01) (coef_ did not converge)
- Base R-Learner class and feeding in XGBoost: -0.03 (-0.04, -0.02) (coef_ did not converge)
- Base X-Learner class and feeding in Linear Regression: -0.26 (-0.45, -0.06) (coef_ did not converge)
- Base X-Learner class and feeding in XGBoost: -0.05 (-0.09, -0.01) (coef_ did not converge)
- Neural Network (MLP): 0.18 (0.08, 0.28)

**CausalML — Cold Months**
- Ready to use S-Learner using Linear Regression: -0.05 (-0.16, 0.05)
- Ready to use T-Learner using XGBoost: -0.03 (-0.08, 0.01)
- Base T-Learner class and feeding in Linear Regression: -0.13 (-0.29, 0.02)
- Base T-Learner class and feeding in XGBoost: -0.03 (-0.08, 0.01)
- Base R-Learner class and feeding in Linear Regression: 0.02 (0.01, 0.03) (coef_ did not converge)
- Base R-Learner class and feeding in XGBoost: 0.01 (0.00, 0.02) (coef_ did not converge)
- Base X-Learner class and feeding in Linear Regression: -0.14 (-0.29, 0.02) (coef_ did not converge)
- Base X-Learner class and feeding in XGBoost: -0.03 (-0.06, -0.00) (coef_ did not converge)
- Neural Network (MLP): 0.18 (0.10, 0.27)

**CausalML — Warm Months**
- Ready to use S-Learner using Linear Regression: -0.07 (-0.17, 0.02)
- Ready to use T-Learner using XGBoost: -0.06 (-0.11, -0.02)
- Base T-Learner class and feeding in Linear Regression: -0.09 (-0.24, 0.05)
- Base T-Learner class and feeding in XGBoost: -0.06 (-0.11, -0.02)
- Base R-Learner class and feeding in Linear Regression: -0.07 (-0.08, -0.06) (coef_ did not converge)
- Base R-Learner class and feeding in XGBoost: -0.04 (-0.05, -0.03) (coef_ did not converge)
- Base X-Learner class and feeding in Linear Regression: -0.10 (-0.24, 0.04) (coef_ did not converge)
- Base X-Learner class and feeding in XGBoost: -0.08 (-0.11, -0.05) (coef_ did not converge)
- Neural Network (MLP): 0.18 (0.10, 0.27)

### Estimate Refutation: Linear Regression

**Full Year**
- Random Common Cause: - p value: -
- Add Unobserved Common Cause: -
- Placebo Treatment Refuter: - p value: -
- Data Subset Refuter: - p value: -

**Cold Months**
- Random Common Cause: - p value: -
- Add Unobserved Common Cause: -
- Placebo Treatment Refuter: - p value: -
- Data Subset Refuter: - p value: -

**Warm Months**
- Random Common Cause: - p value: -
- Add Unobserved Common Cause: -
- Placebo Treatment Refuter: - p value: -
- Data Subset Refuter: - p value: -

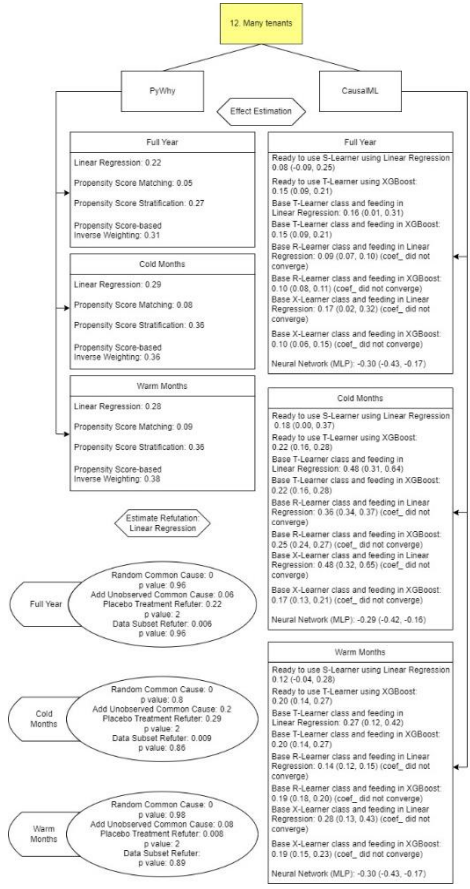
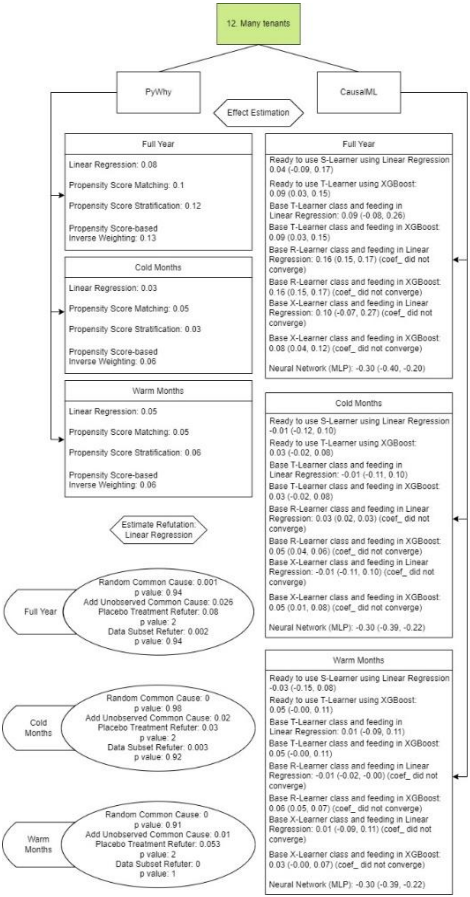
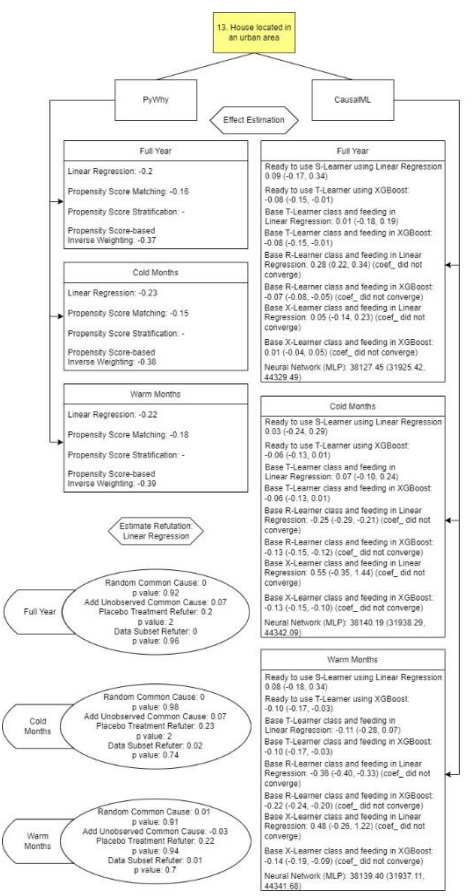
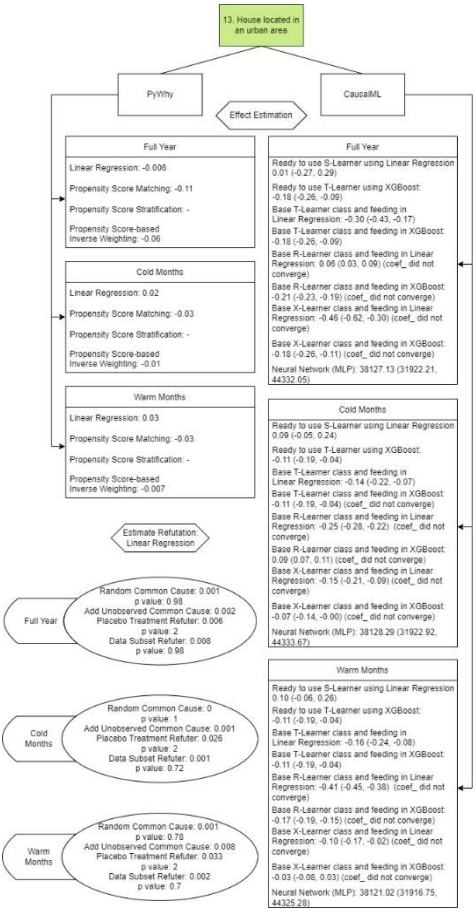

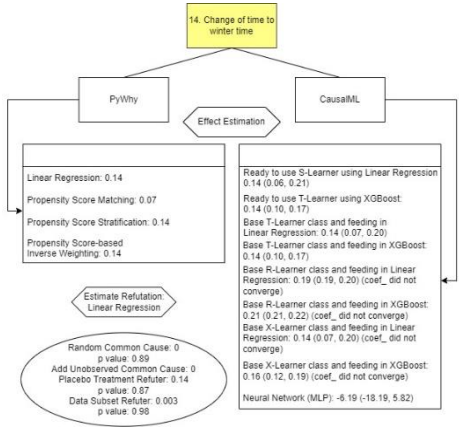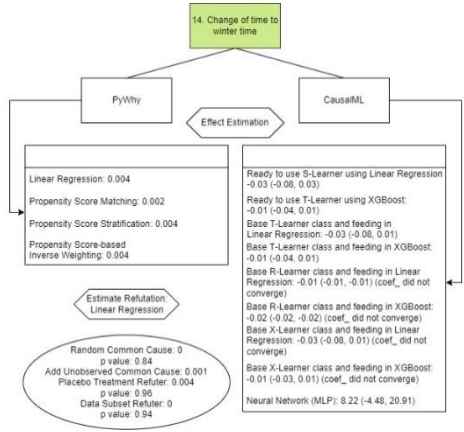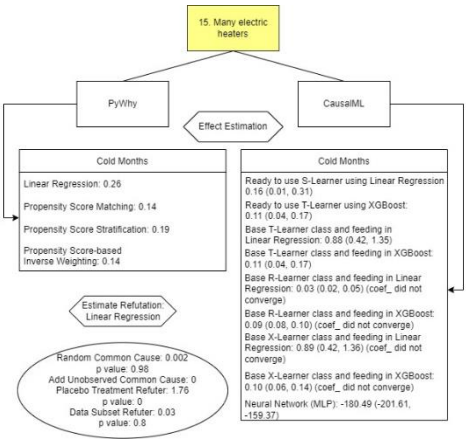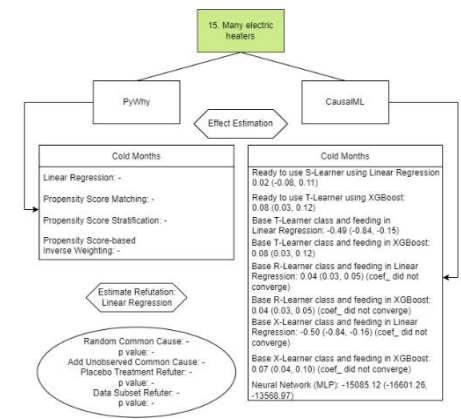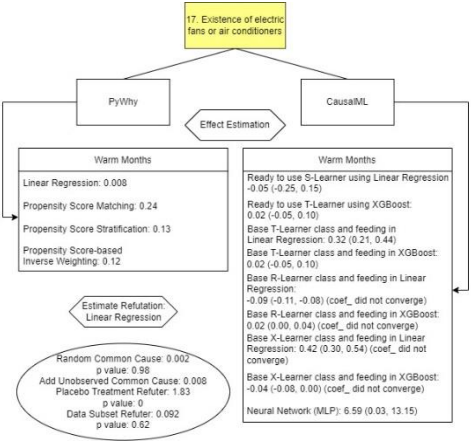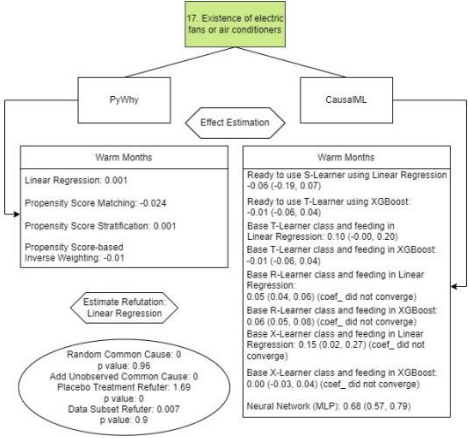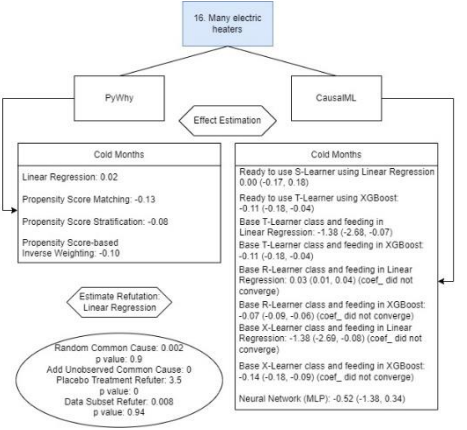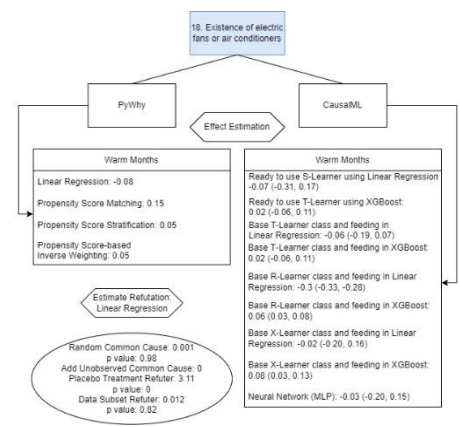